\documentclass[review, 1p]{elsarticle}

\usepackage{lineno,hyperref}
\modulolinenumbers[5]

\journal{Journal of \LaTeX\ Templates}

\usepackage{url}
\usepackage[utf8]{inputenc}
\usepackage{tabularx}

\usepackage{orcidlink}
\usepackage{graphicx}
\usepackage{dcolumn}
\usepackage{bm}
\usepackage[utf8]{inputenc}
\usepackage[T1]{fontenc}
\usepackage{mathptmx}
\usepackage{etoolbox}
\usepackage{xcolor}
\usepackage{graphicx}
\usepackage{lineno,hyperref}
\usepackage{blindtext}
\usepackage{epsfig}
\usepackage{latexsym}
\usepackage{graphics}
\usepackage{epsfig}
\usepackage{amsmath,amssymb,amsfonts,theorem,color,bm}
\usepackage{multirow, multicol, boldline}
\usepackage{booktabs}
\usepackage{stfloats} 
\usepackage{float}
\usepackage{subfig}
\usepackage{soul}
\usepackage{caption}
\hypersetup{colorlinks,allcolors=black}
\usepackage[normalem]{ulem}

\DeclareMathAlphabet\mathbfcal{OMS}{cmsy}{b}{n}

\begin{document}

\begin{frontmatter}

\title{CardioMOD-Net: A Modal Decomposition–Neural Network Framework for Diagnosis and Prognosis of HFpEF from Echocardiography Cine Loops}

\newcommand{\orcidABN}{0000-0002-8539-5405}
\newcommand{\orcidJGM}{0000-0002-7422-5320}
\newcommand{\orcidAA}{0000-0001-6407-6918}
\newcommand{\orcidSLCM}{0000-0003-3605-7351}
\newcommand{\orcidMVO}{0000-0002-7680-3980}
\newcommand{\orcidELP}{0000-0002-2743-1033}


\author[addressUPM]{A. Bell-Navas \orcidlink{\orcidABN}}
\ead{a.bell@upm.es}

\author[addressUPM,addressCCS]{J. Garicano-Mena \orcidlink{\orcidJGM}}
\ead{jesus.garicano.mena@upm.es}

\author[addressCNIC]{A. Ausiello \orcidlink{\orcidAA}}
\ead{antonella.ausiello@cnic.es}

\author[addressUPM]{S. Le Clainche \orcidlink{\orcidSLCM}}
\ead{soledad.leclainche@upm.es}

\author[addressUCM]{M. Villalba-Orero \orcidlink{\orcidMVO}}
\ead{mvorero@ucm.es}

\author[addressCNIC,addressMPCIP]{E. Lara-Pezzi \orcidlink{\orcidELP}}
\cortext[mycorrespondingauthor]{Corresponding author}
\ead{elara@cnic.es}

\address[addressUPM]{ETSI Aeron\'autica y del Espacio - Universidad Polit\'ecnica de Madrid, 28040 Madrid, Spain}

\address[addressCCS]{Center for Computational Simulation (CCS), 28660 Boadilla del Monte, Spain}

\address[addressCNIC]{Centro Nacional de Investigaciones Cardiovasculares Carlos III (CNIC), 28029 Madrid, Spain}

\address[addressUCM]{Facultad de Veterinaria, Universidad Complutense de Madrid, 28040 Madrid, Spain}

\address[addressMPCIP]{Centro de investigación Biomédica en Red Cardiovascular (CIBERCV), 28029 Madrid, Spain}



\begin{abstract}

\textbf{Introduction:} Heart failure with preserved ejection fraction (HFpEF) arises from diverse comorbidities and progresses through prolonged subclinical stages, making early diagnosis and prognosis difficult. Current echocardiography-based Artificial Intelligence (AI) models focus primarily on binary HFpEF detection in humans and do not provide comorbidity-specific phenotyping or temporal estimates of disease progression towards decompensation. We aimed to develop a unified AI framework, CardioMOD-Net, to perform  multiclass diagnosis and continuous prediction of HFpEF onset directly from standard echocardiography cine loops in preclinical models.

\textbf{Methods:} Mouse echocardiography videos from four groups were used: control (CTL), hyperglycaemic (HG), obesity (OB), and systemic arterial hypertension (SAH). Two-dimensional parasternal long-axis cine loops were decomposed using Higher Order Dynamic Mode Decomposition (HODMD) to extract temporal features for downstream analysis. A shared latent representation supported Vision Transformers, one for a classifier for diagnosis and another for a regression module for predicting the age at HFpEF onset.

\textbf{Results:} Overall diagnostic accuracy across the four groups was 65\%, with all classes exceeding 50\% accuracy. Misclassifications primarily reflected early-stage overlap between OB or SAH and CTL. The prognostic module achieved a root-mean-square error of 21.72 weeks for time-to-HFpEF prediction, with OB and SAH showing the most accurate estimates. Predicted HFpEF onset closely matched true distributions in all groups.

\textbf{Discussion:} This unified framework demonstrates that multiclass phenotyping and continuous HFpEF onset prediction can be obtained from a single cine loop, even under small-data conditions. The approach offers a foundation for integrating diagnostic and prognostic modelling in preclinical HFpEF research.

\end{abstract}


\end{frontmatter}

\section{\label{sec:Introduction} Introduction}

Heart failure with preserved ejection fraction (HFpEF) is a major and expanding global health problem, now accounting for more than half of all heart-failure cases and rising steadily in ageing populations \cite{borlaug2020evaluation}. Its clinical complexity derives from the interplay of multiple comorbidities, such as obesity, systemic arterial hypertension, or chronic hyperglycaemia, which frequently coexist and shape distinct patterns of myocardial remodelling. As a consequence, HFpEF does not represent a single disease entity but a spectrum of comorbidity-driven phenotypes that evolve through prolonged subclinical stages before culminating in symptomatic congestion. This heterogeneity complicates both experimental modelling and clinical decision-making and remains a major barrier to identifying early disease or anticipating the transition to symptomatic congestion \cite{borlaug2020evaluation} \cite{akerman2025external}.
Diagnosis relies primarily on echocardiography, integrating morphological, functional, and Doppler parameters. However, the early stages of HFpEF often present with subtle or overlapping phenotypes, particularly across comorbidity-driven subtypes such as obesity, systemic arterial hypertension, or chronic hyperglycaemia. This complexity reduces the sensitivity of conventional diagnostic approaches and frequently results in indeterminate assessments, even when established diagnostic scores are applied to standard echocardiographic images \cite{cassianni2024automated} \cite{akerman2023automated}.

An even greater unmet need lies in prognosis. Clinicians currently lack tools capable of estimating when a patient with risk factors or early HFpEF will progress to decompensation, and even advanced diagnostic algorithms offer limited temporal prognostic precision \cite{akerman2025external} \cite{cassianni2024automated}. Current prognostic models typically provide risk categories rather than specific timelines, and their performance is hindered by the scarcity of longitudinal imaging data.

Artificial Intelligence (AI) offers the potential to capture subtle features of cardiac motion that are not accessible to visual assessment. Yet most echocardiography-based AI studies focus on narrow diagnostic tasks, rely on very large datasets or pre-measured echocardiographic parameters, and treat diagnosis and prognosis as separate problems rather than components of a unified workflow \cite{cassianni2024automated} \cite{akerman2023automated} \cite{kobayashi2022machine}. There is currently no unified framework capable of performing multiclass cardiac phenotyping and continuous prediction of HFpEF onset from standard echocardiographic cine loops.

In this study, we present CardioMOD-Net, an AI tool that addresses these limitations by performing multiclass cardiac phenotyping and continuous prediction of HFpEF onset directly from 2D parasternal long-axis cine loops, within a unified architecture designed to operate effectively in small-data settings (Fig. \ref{fig:results}A).

\section{\label{sec:Methods} Methods}

\subsection{\label{subsec:acq} Data acquisition and experimental groups}

Echocardiography videos were obtained from 37 C57BL/6 mice across four experimental groups: control (CTL), hyperglycaemic (HG), obesity (OB), and systemic arterial hypertension (SAH), as we previously described \cite{villalba2025unraveling}. All procedures followed approved institutional animal care protocols (PROEX177/17) and ARRIVE guidelines. Standard two-dimensional parasternal long-axis (PLAX) cine loops were acquired using a high-frequency ultrasound system (Vevo 2100, 40-MHz probe) \cite{villalba2022non}. Videos had a mean resolution of $692 \times 599$ pixels and contained on average $284 \pm 44$ frames per sequence. For each mouse, multiple echocardiography sequences were collected at different ages. HFpEF onset was defined according to established criteria: pulmonary congestion, measure by lung ultrasound, together with echocardiographic signs of diastolic dysfunction in the presence of preserved left ventricular ejection fraction, and recorded as the age (in weeks) at which each mouse first met this threshold \cite{villalba2025unraveling}.
\subsection{\label{subsec:str} Dataset structure and train–test separation} 
The full dataset comprised 193 echocardiography videos. For each experimental group, approximately 80\% of the sequences were used to train the tool (156 sequences): specifically, 60\% of the sequences were allocated to the training set (118 sequences) and 20\% the validation set (38 sequences). The remaining 20\% were assigned to the test set (37 sequences; Fig. \ref{fig:results}B). To prevent animal-level data leakage, all sequences from a given mouse were included exclusively in either training or testing, ensuring that model evaluation was performed on mice not used during training.

\subsection{\label{subsec:ext} AI framework and feature extraction} 

The unified tool, CardioMOD-Net, integrates two previously published systems for diagnosis and prognosis into a single architecture \cite{groun2025eigenhearts} \cite{bell2025automatic}. Each cine loop was processed using Higher Order Dynamic Mode Decomposition (HODMD) to extract spatiotemporal modes that encode cardiac motion patterns \cite{groun2022higher}. These HODMD-derived features were used to build a shared latent representation supporting both diagnostic and prognostic tasks.
\subsection{\label{subsec:diag} Diagnosis module} 

Heart-state classification (CTL, HG, OB, SAH) was performed using a lightweight Vision Transformer incorporating Masked Autoencoder (MAE)–based strategies \cite{das2024limited} trained on the HODMD feature representation. This design reduces dependence on large datasets and mitigates overfitting by exploiting temporal coherence within cine loops. Diagnostic performance was evaluated using class-specific accuracy and a four-class confusion matrix.

\subsection{\label{subsec:prog} Prognosis module} 

The prognosis component predicted the time-to-HFpEF onset directly from each video. A regression head also based on MAE was trained on the same shared representation used for diagnosis. Prognostic accuracy was quantified using the root-mean-square error (RMSE) between predicted and real HFpEF times for each experimental group.

\subsection{\label{subsec:real} Real-time inference} 

The full pipeline was designed for real-time operation, processing each frame in under 1 s and generating diagnosis and prognosis simultaneously from the same inference pass.

\section{\label{sec:Results} Results}

A total of 193 echocardiography video sequences from 37 mice across four experimental groups were analysed. HFpEF onset times differed markedly between groups, ranging from early events in hyperglycaemic mice ($44.3 \pm 6.7$ weeks) to much later events in controls ($103.8 \pm 22.7$ weeks), reflecting distinct disease trajectories.

\subsection{\label{subsec:diagresults} Diagnosis performance} 

\begin{figure*}[!ht]

\centering
\includegraphics[width=14 cm]{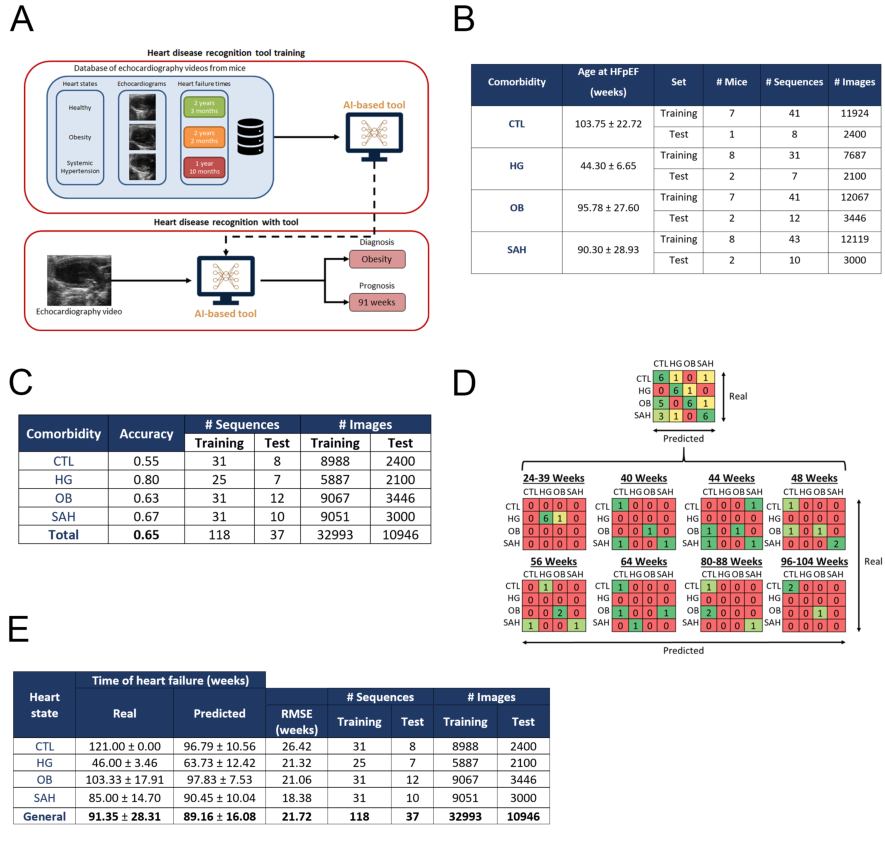}
\caption{\textbf{Diagnosis and prognosis results obtained with the CardioMOD-Net tool (Cardiovascular Modal Decomposition - Neural Network). A}, Schematic depicting the workflow used by this AI-based tool for heart disease recognition. Top, training of the tool for diagnosis and prognosis; bottom, diagnosis and prognosis performed by the tool after it is trained. \textbf{B}, Summary of the main characteristics of echocardiography images in our database. \textbf{C}, Diagnosis results obtained with the tool using the test sequences. \textbf{D}, Confusion matrix obtained with the developed tool using test sequences broken down into age intervals. Top, global confusion matrix; bottom, confusion matrices by age intervals. \textbf{E}, Prognosis results obtained with the tool using the test sequences. CTL, control; HG, hyperglycaemic; OB, obesity; SAH, systemic arterial hypertension; RMSE, Root Mean Squared Error.}
\label{fig:results}

\end{figure*}

The tool achieved an overall diagnostic accuracy of 65\% across the four cardiac conditions (Fig. \ref{fig:results}C). All groups reached accuracies above 50\%, with the best performance in HG mice (80\%), followed by SAH (67\%), OB (63\%), and CTL (55\%). The corresponding confusion matrix (Fig. \ref{fig:results}D) shows that most correct classifications lie on the diagonal, while misclassifications primarily occurred between OB or SH mice and CTL, particularly at younger ages (24–64 weeks). This pattern likely reflects biological overlap during early-stage remodelling and a known limitation of multiclass phenotyping from standard B-mode echocardiography. Despite the modest size of the training set, the model demonstrated the potential of the tool for the diagnosis of cardiac pathologies, especially considering that four different conditions were covered simultaneously in the training process, in contrast to the usual trend, which evaluates single pathology or a binary setting at a time, thereby limiting the potential for clinical application.

\subsection{\label{subsec:progresults} Prognostic performance} 

The unified tool also estimated the time-to-HFpEF onset directly from each video. Across the test set, the model achieved a root-mean-square error (RMSE) of 21.72 weeks (Fig. \ref{fig:results}E). Predicted HF times showed group-specific patterns consistent with data availability and disease trajectories: OB and SAH mice exhibited the most accurate predictions (RMSE 21.06 and 18.38 weeks, respectively), whereas HG mice, which display rapid disease progression and have fewer characteristic frames for learning, showed larger errors (RMSE 21.32 weeks). CTL mice, with later and more variable HFpEF onset, also showed higher deviation (RMSE 26.42 weeks). Importantly, the predicted mean HF times closely matched the real distribution in most groups. For instance, OB mice showed real vs. predicted values of $103.33 \pm 17.91$ weeks and $97.83 \pm 7.53$ weeks, respectively. This suggests that the tool can capture coarse temporal dynamics of ventricular deterioration using only raw cine images, without access to Doppler indices, comorbidity labels, or clinical metadata.

\subsection{\label{subsec:intresults} Integrated task performance} 

Because diagnosis and prognosis are generated from the same shared feature representation, the tool maintains performance in both tasks, showing that effective multiclass diagnosis and prognosis can be achieved even with limited training data. The ability to provide multiclass diagnosis and continuous prognosis simultaneously, in real time and from a single echocardiographic video, represents a practical advantage over traditional AI models, which typically address these tasks separately and require substantially larger datasets.

\section{\label{sec:Discussion} Discussion}

In this study, we present CardioMOD-Net, an AI tool capable of performing multiclass cardiac diagnosis and continuous prediction of HFpEF onset directly from standard PLAX cine loops, addressing methodological gaps that remain unfilled in current echocardiography-based AI. Although recent advances have demonstrated that deep learning can identify HFpEF from single apical four-chamber videos in humans, these approaches remain largely restricted to binary classification and do not provide temporal estimates of disease progression \cite{cassianni2024automated} \cite{akerman2023automated}. Likewise, large multi-task echocardiographic systems trained on millions of clinical images offer comprehensive diagnostic outputs but require extensive datasets and do not address comorbidity-specific phenotyping or time-to-event prediction \cite{akerman2025external}. Our findings show that both diagnosis and prognosis can be extracted from a unified feature representation even under small-data conditions, supporting the feasibility of efficient, task-integrated AI models.

Diagnostic performance was stable across comorbidity-driven phenotypes despite the limited training material, and the prognostic module captured meaningful differences in HFpEF trajectories, particularly in obesity and systemic hypertension. These results suggest that subtle spatiotemporal signatures of ventricular remodelling can be leveraged to approximate disease stage and future deterioration without relying on Doppler indices, biomarkers, or clinical metadata.

The study is limited by its preclinical scope, modest sample size, and reliance on a single echocardiographic view. Nonetheless, the approach offers a foundation for future work aimed at linking comorbidity-specific myocardial dynamics to clinical risk prediction. Integrating similar unified architectures into larger animal cohorts or early-stage human studies may help bridge mechanistic insights and clinically actionable prognostic assessment in HFpEF.

\bibliography{biblio}

\newpage

\section*{\label{sec:Figure} Figure}

\end{document}